\pdfoutput=1
\documentclass[11pt]{article}

\usepackage[preprint]{acl}

\usepackage{bm}
\usepackage{times}
\usepackage{latexsym}
\PassOptionsToPackage{numbers, compress}{natbib}
\usepackage[T1]{fontenc}
\usepackage[utf8]{inputenc}

\usepackage{microtype}

\usepackage{inconsolata}
\usepackage{graphicx}
\usepackage{mdframed}
\usepackage{enumitem}
\usepackage{booktabs}
\usepackage{multirow}
\usepackage{array}
\usepackage{wrapfig,lipsum,booktabs}
\usepackage{multirow}
\usepackage[utf8]{inputenc} % allow utf-8 input
\usepackage[T1]{fontenc}    % use 8-bit T1 fonts
\usepackage{hyperref}       % hyperlinks
\usepackage{url}            % simple URL typesetting
\usepackage{booktabs}       % professional-quality tables
\usepackage{amsfonts}       % blackboard math symbols
\usepackage{nicefrac}       % compact symbols for 1/2, etc.
\usepackage{microtype}      % microtypography
\usepackage{xcolor}         % colors
\usepackage{amsmath,amsfonts}
\usepackage{algorithm}
\usepackage{algorithmic}
\usepackage{float}
\usepackage{xcolor}

\title{QuZO: Quantized Zeroth-Order Fine-Tuning for Large Language Models}
\author{
    Jiajun Zhou$^{1,3}$\footnotemark[2]  Yifan Yang$^{1}$, Kai Zhen$^{2}$, Ziyue Liu$^{1}$, Yequan Zhao$^{1}$,\\
    \textbf{Ershad Banijamali}$^{2}$, \textbf{Athanasios Mouchtaris}$^{2}$,  \textbf{Ngai Wong}$^{3}$, \textbf{Zheng Zhang}$^{1}$ \\
  $^1$University of California, Santa Barbara, $^2$Amazon AGI \\
  $^3$The University of Hong Kong\\
  \texttt{\{jjzhou,nwong\}@eee.hku.hk}, \texttt{zhengzhang@ece.ucsb.edu} }

\begin{document}
\maketitle
\renewcommand{\thefootnote}{\fnsymbol{footnote}}
\footnotetext[2]{Work undertaken during the visit at UC Santa Barbara}
% \footnotetext[2]{Work undertaken during the visit at UC Santa Barbara}
% \DeclareMathAlphabet\mathbfcal{OMS}{cmsy}{b}{n}
\newcommand{\ten}[1]{\mathbfcal{#1}}
\newcommand{\mat}[1]{\mathbf{#1}}

\begin{abstract}

Language Models (LLMs) are often quantized to lower precision to reduce the memory cost and latency in inference. However, quantization often degrades model performance, thus fine-tuning is required for various down-stream tasks. Traditional fine-tuning methods such as stochastic gradient descent and Adam optimization require backpropagation, which are error-prone in the low-precision settings. To overcome these limitations, we propose the Quantized Zeroth-Order (QuZO) framework, specifically designed for fine-tuning LLMs through low-precision (e.g., 4- or 8-bit) forward passes. Our method can avoid the error-prone low-precision straight-through estimator, and utilizes optimized stochastic rounding to mitigate the increased bias. QuZO simplifies the training process, while achieving results comparable to first-order methods in ${\rm FP}8$ and superior accuracy in ${\rm INT}8$ and ${\rm INT}4$ training. Experiments demonstrate that low-bit training QuZO achieves performance comparable to MeZO optimization on GLUE, Multi-Choice, and Generation tasks, while reducing memory cost by $2.94 \times$ in LLaMA2-7B fine-tuning compared to quantized first-order methods.

% Experiments demonstrate that QuZO significantly reduces memory consumption by a factor of 5 in LLaMa-7B fine-tuning, compared to first-order optimization with the Fully Sharded Data Parallel (FSDP) engine, when employing an INT8 quantization scheme.
\end{abstract}

\section{Introduction}

Large Language Models (LLMs) have achieved state-of-the-art performance in natural language processing, impacting various science and engineering fields. However, deploying and fine-tuning LLMs consumes significant hardware resources because of their huge model size. To address this issue, extensive research has focused on LLM quantization~\citep{brown2020language,yuan2024llm}. Notable approaches include post-training quantization~\citep{yao2022zeroquant,wu2023zeroquant}, quantization-aware training~\citep{bhalgat2020lsq+,liu2023llm,nagel2021white}, and fully quantized training~\citep{choukroun2019low,xi2023training,markidis2018nvidia}. Post-training quantization can effectively reduce the latency and memory costs of inference, but often leads to a significant accuracy drop in low-precision formats, although various techniques~\citep{shao2023omniquant,xiao2023smoothquant,lin2023awq,liu2023llm} can partially mitigate this issue.  Quantization-aware training~\cite{liu2023qllm} offers better accuracy, but is more expensive due to the use of high-precision computational graphs. Truly quantized training methods employ low-precision gradients, activation, and weights to reduce hardware costs~\cite{wang2018training,banner2018scalable,micikevicius2017mixed}. However, implementing truly quantized training requires advanced hardware and software support for both forward and backward passes. Meanwhile, the straight-through estimator~\cite{yin2019understanding}, which is commonly used for quantized gradient estimations,  often causes unstable and inaccurate results in low-bit training. 

In practice, LLM users may afford only a low-cost LLM inference engine (e.g., an edge FPGA or embedded system) with limited precision (e.g., ${\rm INT}8$ or ${\rm INT}4$). This paper asks the following question: {\it Can we leverage inference-only quantized hardware to fine-tune low-bit LLMs while achieving good performance?} This seems challenging because (1) inference-only hardware lacks sufficient memory bandwidth and storage to retain intermediate activations required for backpropagation, and (2) the Straight-Through Estimator (STE) introduces increasing gradient approximation errors in lower-bit formats~\citep{malinovskii2024pv}. 

\begin{figure}
         \centering
         % \vspace{-15pt}
         % \includegraphics[width=2.8in]{figure/figure1.pdf}
         \includegraphics[width=3.in]{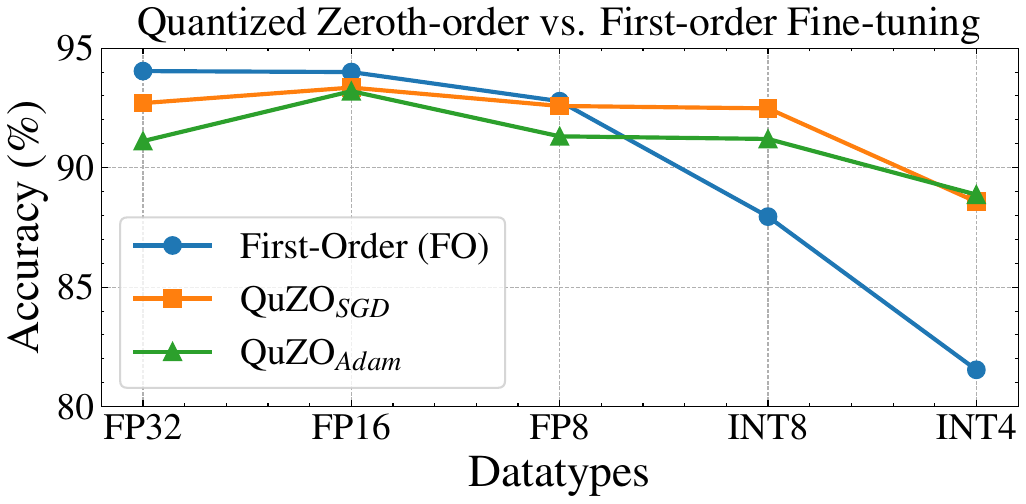}
         % \vspace{-20pt}
   \caption{The proposed QuZO provides higher fine-tuning accuracy than first-order (FO) methods in ultra-low precision on the RoBERTa-Large model.}
    \label{fig:enter-label}
    \vspace{-5pt}
\end{figure}

The recent MeZO~\citep{malladi2024fine} enables memory-efficient zeroth-order (ZO) fine-tuning for LLMs, but suffers from an avoidable performance drop compared to first-order (FO) methods due to the bias and variance of ZO gradient estimation. In this paper, we show that {\it a quantized zeroth-order optimizer (QuZO) can achieve better accuracy than its first-order counterparts in a low-precision setting}. Fig.~\ref{fig:enter-label} shows that both the QuZO and FO methods experience accuracy drops as the quantization precision decreases, which is well expected. However, QuZO consistently outperforms FO methods when the quantization precision is ${\rm INT}8$ or below. Unlike traditional FO quantized training that depends on error-prone STE~\citep{yin2019understanding}, our QuZO optimizer is more resistant to quantization error. Our contributions are summarized below. 
\begin{itemize}[leftmargin=*]
\vspace{-8.1pt}

\item We identify the challenge of naive quantized ZO training, and propose a stochastic quantized perturbation method with theoretical soundness to reduce bias in quantized ZO gradient estimation. %This approach is more resilient in low-bit fine-tuning.

\vspace{-8.1pt}

\item We introduce the implementation of QuZO as a plugin that integrates seamlessly with a quantized LLM inference engine, enabling accurate fine-tuning of low-bit LMs without backpropagation.

\vspace{-8.1pt}
\item We provide detailed numerical analysis about the proposed gradient estimator and the QuZO training framework. We show the benefit of our quantized ZO gradient estimator and the better training behavior of QuZO in low-bit LLM fine-tuning (especially ${\rm INT}4$-format trainig).

\vspace{-8.1pt}
\item We apply QuZO to fine-tune 4/8-bit LLMs using both full-model fine-tuning and Low-Rank Adaptation (LoRA). QuZO achieves much better accuracy than quantized first-order training while reducing the  memory cost by $1.4\times-2.94\times$.
% by $1.4-5.46\times$ \zz{Make sure that the data is updated.} compared to traditional quantization-aware \zz{quantization-aware or truly quantized?} training.

\end{itemize}

\section{Related Work}
\label{gen_inst}
\paragraph{Zeroth-order method.} Zero-order (ZO) optimization techniques use forward passes for gradient estimation. Since backpropagation is not required during training, ZO methods reduce memory consumption significantly compared to a FO method. MeZO~\citep{malladi2024fine} employed a memory-efficient ZO stochastic gradient descent (ZO-SGD) algorithm to efficiently fine-tune LLMs exceeding $60$ billion parameters, leveraging parameter-efficient approaches~\cite{yang2024loretta,liu2021p} like LoRA~\cite{hu2021lora}. Other ZO methods include ZO-SGD~\citep{ghadimi2013stochastic} and ZO-Sign-SGD~\citep{liu2018signsgd} using sign-based gradient estimation, the ZO-Adam~\citep{chen2019zo} optimizer exploiting momentum information, and parameter-efficient methods like AdaZeta \cite{yang2024adazeta}. Sparse MeZO~\cite{liu2024sparse} employs a sparse perturbation for LLM fine-tuning. ${\rm FP}16$ ZO training~\citep{zhang2024revisiting} performs well but still faces memory bottlenecks. Recent ZO quantization introduces fixed-point 16-bit but fails at 8-bit~\cite{feng2024stepping}. However, we overcome the challenges of lower-precision quantization and enable accurate fine-tuning of LLMs below 8-bit quantization.

% However, no prior ZO work has explored low-bit quantization such as 8-bit for hardware deployment. \zz{This is NOT true. I remember that Sijia Liu's benchmark paper includes some low-precision examples. Some recent work also utilized quantized ZO training. Please cite these papers and explain the difference with QuZO.}

\paragraph{Quantization of LLMs.} Various quantization methods have been developed to reduce the memory and computing cost of LLMs. LLM.int8()~\citep{dettmers2022gpt3} reduces the precision of model weights while keeping outliers in ${\rm FP}16$. SmoothQuant \citep{xiao2023smoothquant} introduces a fine-grained quantization method that supports ${\rm INT}8$ operations exclusively. QLLM~\citep{liu2023qllm} addresses the outlier problem via employing an adaptive channel reassembly technique. LLM-QAT~\citep{liu2023llm} employs quantization-aware training with a data-free strategy to achieve $4$-bit quantization. Furthermore, the $4$-bit training~\citep{xi2023training} and QLoRA~\citep{dettmers2024qlora} methods leverage a Hadamard Transform and a novel NF4 datatype, respectively, to accelerate training while preserving performance. While prior quantized training methods rely on backpropagation for gradient updates, our QuZO method eliminates the error-prune STE-based back propagation and used low-bit inference for truly quantized fine-tuning.

% \section{Preliminaries}
% \label{headings}

\begin{figure*}[h]
    \centering
    \includegraphics[width=1\linewidth]{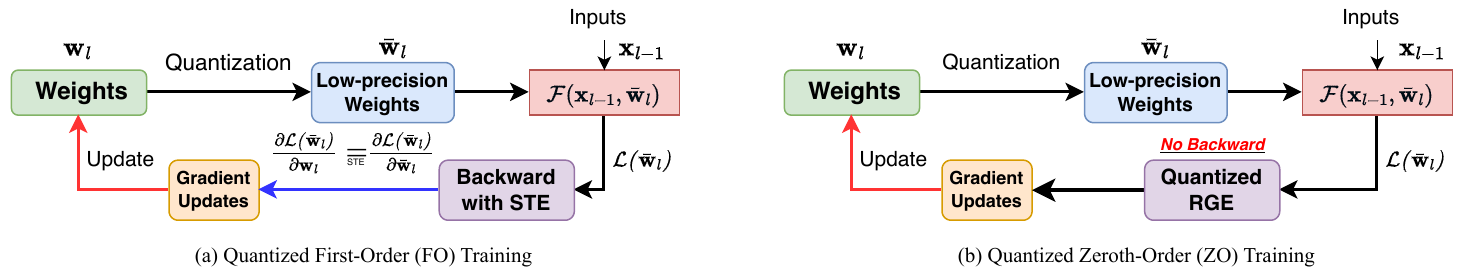}
    \caption{Computational graphs for quantized first-order (FO) and zeroth-order (ZO) training.}
    \label{fig:qzo_framework}
    \vspace{-10pt}
\end{figure*}

\section{The QuZO Fine-Tuning Method}
%In this section, we present QuZO for fine-tuning large language models (LLMs) with low precision.  
We start with a high-level introduction to our QuZO framework. Given a quantized LLM inference model, QuZO uses a low-bit ZO optimizer to update quantized model parameters directly during training. We assume that the forward pass $ \mat{x}_l=\mathcal{F}(\mat{x}_{l-1}, \bar{\mat{w}}_{l})$ computes the output of the $l$-th layer using the quantized weight matrix $\bar{\mat{w}}_{l}$ and the previous-layer feature \(\mat{x}_{l-1}\), as shown in Fig.~\ref{fig:qzo_framework} (b). %Here, \(\mat{X}^l \in \mathbb{R}^{n \times d_l}\) represents the activations for $n$ samples, and $\mat{W}^l \in \mathbb{R}^{d_{l-1} \times d_l}$ is the model parameters in layer $l$.
% where \(\textbf{X} \in \mathbb{R}^{N \times D}\) is a matrix representing a batch of data, N is the batch size and D is the feature dimension. To optimize the weight parameter \(\textbf{W}\), we utilize the loss function, such as stochastic gradient descent (SGD) \cite{amari1993backpropagation}. In our QuZO algorithm, the update is expressed as \( w_{t+1} = w_{t} -  {n_{t}} * {\rm projected}_{\rm grad} * z_{i} \) as shown in Algorithm \ref{alg:quzo}. where \(\mat{X} \in \mathbb{R}^{N \times D}\) is a matrix representing a batch of data, $N$ is the batch size and $D$ is the feature dimension. 
With just a few forward passes, our QuZO framework uses quantized RGE (see Section 3.2) to estimate ZO gradients, eliminating the need for BP in model updates. This approach fundamentally differs from existing quantized training methods shown in FigFig.~\ref{fig:qzo_framework} (a), which uses STE in the BP to approximate quantized gradient  \(\frac{\partial \cal L (\bar{\mat{w}})}{\partial \bar{\mat{w}}_{l}}\). Our method avoids the straight-through estimator (STE)~\cite{yin2019understanding} used in truly quantized FO training, enabling high-accuracy training on a low-precision hardware platform. 

In the following, we first show the challenges of ZO-SGD in the quantized setting, and then propose a solution to address this fundamental challenge.

\subsection{Challenges of Quantized ZO Training}
Standard ZO-SGD uses a randomized gradient estimator (RGE)~\cite{nesterov2017random,ghadimi2013stochastic} to approximate a full-precision gradient. Specifically, given full-precision model parameters $\mat{w} \in \mathbb{R}^d$, a loss function ${\cal L}(\mat{w}, \mathcal{B})$ and a minibatch of dataset $\mathcal{B}$, RGE computes the gradient as:
\begin{align}
 \label{equation:mezo}
  % \begin{center} 
\nabla\hat{{\cal L}}(\mat{w}) &=  \sum_{i=1}^{n} \frac{{\cal L}_\mathcal{B}({\mat{w}} + \epsilon \mat{u}_{i}) - {\cal L}_\mathcal{B}({\mat{w}} - \epsilon \mat{u}_{i})}{2n\epsilon} \mat{u}_{i} \nonumber \\
& \approx \frac{1}{n} \sum \limits_{i=1}^n  \mat{u}_i \mat{u}_i^T \nabla {\cal L}_\mathcal{B}({\mat{w}}),
% \end{center}  
\end{align}
where $\epsilon$ is a scaling factor, $\{\mat{u}_i\}_{i=1}^n$ are i.i.d. samples drawn from certain distributions with a unit variance (e.g., a standard Gaussian distribution).  While $\nabla\hat{{\cal L}}(\mat{w})$ differs from the true gradient $\nabla {\cal L}_\mathcal{B}({\mat{w}})$, its expectation serves as a good gradient estimator because
\begin{align}
 \label{equation:mezo_expect}
  % \begin{center} 
\mathbb{E}\left[\nabla\hat{{\cal L}}(\mat{w}) \right] & \approx \frac{1}{n} \sum \limits_{i=1}^n \mathbb{E} \left( \mat{u}_i \mat{u}_i^T \right) \nabla {\cal L}_\mathcal{B}({\mat{w}}) \nonumber \\
&= \nabla {\cal L}_\mathcal{B}({\mat{w}}).
% \end{center}  
\end{align}
This statistical property ensures the asymptotical convergence of ZO-SGD. Assuming the quantized model parameters \(\bar{\mat{w}}\) are available and only low-precision hardware is used for inference, the full-precision random perturbation \(\mat{u}_i\) cannot be directly applied to \(\bar{\mat{w}}\) due to hardware limitations. To address this, \(\mat{u}_i\) is replaced with its quantized counterpart \(\hat{\mat{u}}_i = Q(\mat{u}_i)\), leading to a low-precision RGE: 
\begin{align}
 \label{equation:mezo_quant}
  % \begin{center} 
\nabla\hat{{\cal L}}(\bar{\mat{w}}) &=  \sum_{i=1}^{n} \frac{{\cal L}_\mathcal{B}\left({\bar{\mat{w}}} + \epsilon \hat{\mat{u}}_i \right) - {\cal L}_\mathcal{B}\left ({\bar{\mat{w}}} - \epsilon \hat{\mat{u}}_i  \right)}{2n\epsilon} \hat{\mat{u}}_i \nonumber \\
& \approx \frac{1}{n} \sum \limits_{i=1}^n  \hat{\mat{u}}_i \hat{\mat{u}}_i^T \nabla {\cal L}_\mathcal{B}({\bar{\mat{w}}}).
% \end{center}  
\end{align}
Taking the exception values on both sides, we have
\begin{align}
 \label{equation:mezo_quant_expect}
  % \begin{center} 
\mathbb{E}\left[\nabla\hat{{\cal L}}(\bar{\mat{w}}) \right] & \approx \frac{1}{n} \sum \limits_{i=1}^n \mathbb{E} \left( \hat{\mat{u}}_i \hat{\mat{u}}_i^T \right) \nabla {\cal L}_\mathcal{B}({\bar{\mat{w}}}) \nonumber \\
& \neq \nabla {\cal L}_\mathcal{B}({\bar{\mat{w}}})
% \end{center}  
\end{align}
Since the quantized perturbation \(\hat{\mat{u}}_i = Q(\mat{u}_i)\) no longer maintains a unit variance, the above naive quantized RGE introduces bias during fine-tuning and may lead to divergence in training. 
\subsection{Proposed Quantized RGE}
We propose a new quantized RGE scheme to address the challenge in the previous subsection.

\paragraph{Stochastic Quantization of $\mat{u}_i$.} We first define a quantization operation of $Q(\mat{u}_i)$ based on stochastic rounding~\cite{connolly2021stochastic}:
% \begin{equation}
% \label{equation:stochastic_quant}
% \resizebox{0.85\hsize}{!}{$
% Q(\mat{u}_i) = \text{clamp} \Big( \lfloor s_u \mat{u}_i \rfloor + \text{Ber}(s_u \mat{u}_i - \lfloor s_u \mat{u}_i \rfloor), \, L_\text{min}, \, L_\text{max} \Big) + z_0
% $}
% \end{equation}
\begin{align}
\label{equation:stochastic_quant}
&Q(\mat{u}_i) = \text{clamp} \Big(SQ, L_\text{min}, \, L_\text{max} \Big) + z_0 , \notag \\
&SQ = \Big( \lfloor s_u \mat{u}_i \rfloor + \text{Ber}(s_u \mat{u}_i - \lfloor s_u \mat{u}_i \rfloor)
\end{align}
The stochastic quantization formula $Q(\mat{u}_i)$ converts the perturbation $\mat{u}_i$ into a low-bit representation by scaling it with a factor $s_{u}$ as $s_u \mat{u}_i$, performing a downward rounding operation $\lfloor s_u \mat{u}_i \rfloor$, and applying stochastic up-rounding using a Bernoulli random variable $\text{Ber}(s_u \mat{u}_i - \lfloor s_u \mat{u}_i \rfloor)$. The resulting quantized value is clamped to the representable range $[L_\text{min}, L_\text{max}]$ and shifted by the zero point $z_0$. This stochastic rounding ensures that
\begin{equation}
    \mathbb{E}_{Q} \left[ Q(\mat{u}_i)\right]= 
 \mathbb{E}\left[\mat{u}_i\right].
\end{equation} 

We can produce two different quantization results by using two random seeds in the stochastic rounding full-precision $\mat{u}_i$: 
\begin{align}
\label{eq:two_quant_vec}
    \mat{u}_{i,1}& =Q_{1}(\mat{u}_i) = Q(\mat{u}_i) \; \mbox{with random seed} \; i_1 ; \nonumber \\
    \mat{u}_{i,2}& =Q_{2}(\mat{u}_i) = Q(\mat{u}_i) \; \mbox{with random seed} \; i_2; \nonumber \\
    \mat{u}_{i,1} &\neq \mat{u}_{i,2}.
\end{align}
The above stochastic quantizations ensure that (1) the expectation of the quantized perturbations \( \mat{u}_{i,1} \) and \( \mat{u}_{i,2} \) equals the original perturbation \( \mat{u}_i \), (2)  \( \mat{u}{i,1} \) and \( \mat{u}_{i,2} \) are conditionally independent to each other. As a result, we have
% These two quantization results are conditionally independent of each other given $\mat{u}_i$, and thus we have:
\begin{align}
    &\mathbb{E}_{Q_1}(\mat{u}_{i,1})= \mathbb{E}_{Q_2}(\mat{u}_{i,2})=\mat{u}_i, \nonumber \\
   & \mathbb{E}_{Q_1, Q_2}(\mat{u}_{i,1} \mat{u}_{i,2}^T)=\mathbb{E}_{Q_1}(\mat{u}_{i,1}) \mathbb{E}_{Q_2}(\mat{u}_{i,2}^T)= \mat{u}_i \mat{u}_i^T. \nonumber 
\end{align}

\paragraph{Our Quantized RGE.} With the two conditionally independent quantized vectors $\mat{u}_{i,1}$ and $\mat{u}_{i,2}$ defined in Eq.~\eqref{eq:two_quant_vec}, we propose the following quantized RGE: 
\begin{equation}
 \label{equation:qzo}
  % \begin{center} 
  \resizebox{0.87\hsize}{!}{$\nabla\hat{{\cal L}}(\bar{\mat{w}}) =  \sum \limits_{i=1}^{n} \frac{{\cal L}_\mathcal{B}(\bar{\mat{w}} + \epsilon \mat{u}_{i,1}) - {\cal L}_\mathcal{B}(\bar{\mat{w}} - \epsilon \mat{u}_{i,1})}{2n\epsilon} \mat{u}_{i,2}$}
% \end{center}  
\end{equation}
As $\epsilon$ $\to$ 0, the RGE result is
\begin{equation}
 \label{equation:qzo_approx}
  % \begin{center} 
\nabla\hat{{\cal L}}(\bar{\mat{w}}) \approx  \frac{1}{n} \sum \limits_{i=1}^n  {\mat{u}}_{i,1} {\mat{u}}_{i,2}^T \nabla {\cal L}_\mathcal{B}({\bar{\mat{w}}}).
\end{equation}
The estimation results depend on three random vectors and functions: $\mat{u}_i$, $Q_1$ and $Q_2$. Taking expectation values on both sides of Eq.~\eqref{equation:qzo_approx}, we have
\begin{align}
\label{eq:quantized_RGE_expect}
&\mathbb{E}\left[ \nabla\hat{{\cal L}}(\bar{\mat{w}})  \right]  \approx  \frac{1}{n} \sum \limits_{i=1}^n  \mathbb{E}_{\mat{u_i}, Q_1, Q_2}\left[ {\mat{u}}_{i,1} {\mat{u}}_{i,2}^T \right] \nabla {\cal L}_\mathcal{B}({\bar{\mat{w}}}) \nonumber \\
&= \frac{1}{n} \sum \limits_{i=1}^n  \mathbb{E}_{\mat{u_i}} \left[ \mathbb{E}_{ Q_1, Q_2}\left[ {\mat{u}}_{i,1} {\mat{u}}_{i,2}^T \right] \right] \nabla {\cal L}_\mathcal{B}({\bar{\mat{w}}}) \nonumber\\
&=\frac{1}{n} \sum \limits_{i=1}^n \mathbb{E} \left( {\mat{u}}_i {\mat{u}}_i^T \right) \nabla {\cal L}_\mathcal{B}({\bar{\mat{w}}}) \nonumber \\
& = \nabla {\cal L}_\mathcal{B}({\bar{\mat{w}}}). 
\end{align}
The expectation value of our quantized RGE remains a reliable estimator of the true gradient, which is similar to the full-precision RGE. This indicates that our proposed RGE will ensure asymptotical convergence as in a full-precision ZO method. This theoretical property ensures excellent training performance even in low-precision settings (e.g. $\texttt{INT8}$ and $\texttt{INT4}$).

\subsection{Implementation of QuZO}
Now we present the details of the implementation of the QuZO framework.

\begin{algorithm*}[t]
\footnotesize
\renewcommand{\algorithmicrequire}{\textbf{Require:}}
\renewcommand{\algorithmicensure}{\textbf{Output:}}
\caption{QuZO: Quantized Zeroth-Order Training}
\label{alg:quzo}
\begin{algorithmic}[1]
\REQUIRE LLM model parameters ${\mat{w}}$ $\in$ $\mathbb{R}^{d}$, learning rate ${\eta_{t}}$, $T$ is the step, perturbation scaling factor $\epsilon$ and dataset $\mathcal{B}$.
\STATE Initial Pre-trained Model to Quantized Model or directly load a quantized model.
\STATE $\bar{{\mat{w}}} = {\rm Q}({\mat{w}})$ {\hfill $\lhd$ Optionally, quantize the model if starting with a full-precision model}
% \STATE QuZO fine-tuning initiated.
\FOR{t \textbf{in} $T$}
\FOR{i \textbf{in} $n$}
\STATE $ \mat{u}_{i,1} \leftarrow Q_{1}(\mat{u}_i), \mat{u}_i\sim \mathcal{N}(0, \mathbb{I}_d) $ {\hfill $\lhd$ Quantize the perturbation $\mat{u}_i$ with a random seed $i_1$}
\STATE $\mat{u}_{i,2} \leftarrow Q_{2}(\mat{u}_i) $ {\hfill $\lhd$ Quantize the perturbation $\mat{u}_i$ with a random seed $i_2$}
\STATE $\bar{\mat{w}}_{t} \leftarrow \bar{{\mat{w}}}_{t} + \epsilon \cdot \mat{u}_{i,1} $  {\hfill $\lhd$ Low-bit stochastic perturbation updates \(\bar{\mat{w}}_{t}\) using positive scaling}
\STATE ${\cal L}_{1}^{i}  \leftarrow  {\rm \mathcal{F}} (\bar{{\mat{w}}}_{t},\mathcal{B})$  {\hfill $\lhd$ First zeroth-order forward pass}
\STATE $\bar{{\mat{w}}}_{t} \leftarrow \bar{{\mat{w}}}_{t} - 2\epsilon \cdot \mat{u}_{i,1} $  {\hfill $\lhd$ Low-bit stochastic perturbation updates \(\bar{\mat{w}_{t}}\) using negative scaling}
\STATE ${\cal L}_{2}^{i} \leftarrow  {\rm \mathcal{F}}  (\bar{{\mat{w}}}_{t},\mathcal{B})$  {\hfill $\lhd$ Second zeroth-order forward pass}
\STATE $\mu_i \leftarrow    ({\cal L}_{1}^{i} - {\cal L}_{2}^{i})/(2\epsilon) $ {\hfill $\lhd$ Sensitivity w.r.t. the quantized perturbation}
\STATE $\bar{{\mat{w}}}_{t} \leftarrow \bar{{\mat{w}}}_{t} + \epsilon \cdot \mat{u}_{i,1} $ {\hfill $\lhd$ Recover \(\bar{\mat{w}}_{t}\) to its original state}
\STATE $\bar{{\mat{w}}}_{t+1} \leftarrow  \bar{\mat{w}}_{t} - Q( \frac{\eta_t  \mu_i }{n} \mat{u}_{i,2} ) $  {\hfill $\lhd$ Quantized LLM model update}
% \STATE $\mat{P}_g \leftarrow    ({\cal L}_{1}^{i} - {\cal L}_{2}^{i})/(2\epsilon) $ {\hfill $\lhd$ Projected gradient calculation}
% \STATE $\bar{\mat{P}}_g[i] \leftarrow   Q(\mat{P}_g) $ {\hfill $\lhd$ Uniform quantization is applied to quantize \(\mat{P}_g\)}

\ENDFOR

\ENDFOR

\STATE \textbf{return} $\bar{{\mat{w}}}$ {\hfill $\lhd$ Return a quantized model}
\end{algorithmic}
\end{algorithm*}

\paragraph{Quantized Model Updates.} Recall that in full-precision ZO-SGD, the gradient is computed in~\eqref{equation:mezo}, and the model parameters are updated as 
\begin{align}
\label{eq:FP-gradient}
\mat{w}_{t+1} &= \mat{w}_{t} - \eta_{t} \cdot \nabla\hat{{\cal L}}(\mat{w}_{t})
\end{align}
where $\mat{w}_{t}$ represents the model parameters at iteration $t$, $\eta_{t}$ is the learning rate and $\nabla\hat{{\cal L}}(\mat{w}_{t})$ denotes the estimated gradient of the loss function. Since $\mat{w}_t\approx s_w \bar{\mat{w}}_t$, and $s_w$ is a scaling factor used in the quantization $\bar{\mat{w}}_t=Q\left( \mat{w}_t/s_w \right)$, with $Q[\cdot]$ representing the stochastic quantization applied to the parameters. This approximation suggests:
\begin{align}
\label{eq:FP-gradient_approx}
\mat{w}_{t+1} &\approx s_w\left[ \bar{\mat{w}}_{t} - \eta_{t} \cdot \nabla\hat{{\cal L}}(\bar{\mat{w}}_{t}) \right]
\end{align}
To achieve a {\it truly quantized} training process suitable for low-precision hardware, the model parameters are updated as:
% In order to perform {\it truly quantized} training on a resource-constrained low-precision hardware, we can update the quantized model parameters as
\begin{align}
\label{eq:quantized-gradient}
\bar{\mat{w}}_{t+1} &= \bar{\mat{w}}_{t} -Q\left[ \eta_{t} \cdot \nabla\hat{{\cal L}}(\bar{\mat{w}}_{t}) \right].
\end{align}
To refine the update process, multiple steps can be used. For each query $i$, we compute
% Here $Q[\cdot]$ again is a stochastic quantization to avoid bias. We can also perform the update via multiple steps. Specifically, let 
\begin{align}
\label{eq:delta_l}
\mu_i & = \frac{{\cal L}_\mathcal{B}(\bar{\mat{w}} + \epsilon \mat{u}_{i,1}) - {\cal L}_\mathcal{B}(\bar{\mat{w}} - \epsilon \mat{u}_{i,1})}{2\epsilon}.
\end{align}
Then the quantized model $\bar{\mat{W}}$ is updated as
\begin{align}
\label{eq:quantized-gradient-2}
\bar{\mat{w}}_{t+1} &= \bar{\mat{w}}_{t} -\sum \limits_{i=1}^n Q\left( \frac{\eta_t  \mu_i }{n} \mat{u}_{i,2}\right).
\end{align}
Here $\mat{u}_{i,2}$ is a second quantized version of $\mat{u}_i$ as explained in Eq.~\eqref{eq:two_quant_vec}. Stochastic rounding $Q[\cdot]$ ensures that no additional bias will be introduced when we update the LLM parameters directly at low precision. 

\paragraph{Algorithm Flow.} The pseudo codes of QuZO are summarized in Algorithm~\ref{alg:quzo}. For each query $i$, two forward passes are performed to determine the sensitivity ($\mu_i$) of the loss function with respect to a quantized perturbation direction $\mat{u}_{i,1}$ (lines 5-11). The resulting low-precision gradient associated with each inquiry is obtained by quantizing a scaled version of $\mat{u}_{i,2}$, where the sensitivity ($\mu_i$), the learning rate $\eta_t$, and the sample size $n$ are taken into account. This low-precision ZO gradient allows us to directly update the quantized LLM model parameters with low-precision hardware.

\paragraph{QuZO for LoRA.}
We can extend the QuZO framework by incorporating low-rank adaptation to allow low-precision parameter-efficient fine-tuning.  Our approach uses the model quantization strategies of QLoRA~\citep{dettmers2024qlora} and LLM.int8()~\citep{dettmers2022gpt3} without modifying the quantized model. QuZO significantly reduces memory overhead by eliminating the storage of FO optimizer states and updating only the low-rank trainable matrices \(\mat{A} \in \mathbb{R}^{d \times r}\) and \(\mat{B} \in \mathbb{R}^{r \times d}\) using forward passes. In QuZO fine-tuning, the model parameters are quantized and frozen at low precision (e.g. 4 or 8 bits), and we update solely on the low-rank matrices $\mat{A}$ and $\mat{B}$. The trainable low-rank matrices are quantized (denoted as $Q[\mat{A}]$ and $Q[\mat{B}]$) in order to match the precision of the LLM . By doing so QuZO training can significantly further reduce the memory cost compared to traditional LoRA for 4/8-bit LLM fine-tuning. %\zz{How can the memory saving be so large? Have you considered the memory cost by the full model?}%This capability makes QuZO highly suitable for the deployment of quantized LLMs as a service, aligning with the core advantages of PEFT methods.

\subsection{QuZO Analysis}
In this subsection, we analyze the quality of gradient estimation in QuZO and its impact to training.

\begin{figure}[t]
    \centering
    \includegraphics[width=\linewidth]{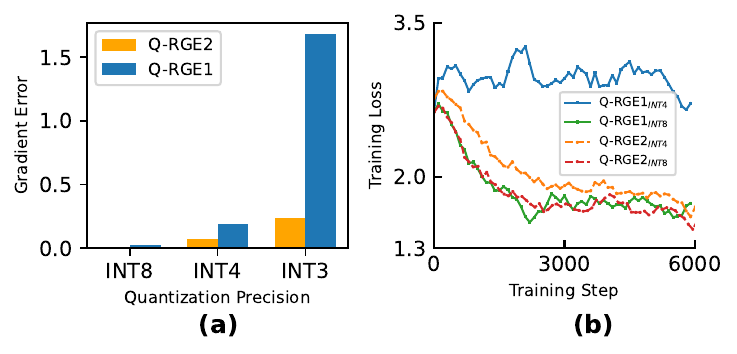}
    \caption{(a) Errors of quantized gradient estimation Q-RGE1 in Eq.~\eqref{equation:mezo_quant} and our proposed Q-RGE2 in Eq.~\eqref{equation:qzo}. (b) Training loss of low-precision ZO optimizer with these two quantized gradient estimators, respectively.}
    \label{fig:quzo_analysis}
    \vspace{-10pt}
\end{figure}

{\bf QuZO Gradient Quality.} We use a simple encoder-block transformer to analyze the asymptotic behavior of two quantized ZO gradient estimators. Q-RGE1 refers to the quantized estimate in Eq.~\eqref{equation:mezo_quant}, and Q-RGE2 denotes our proposed estimation in Eq.~\eqref{equation:qzo}. Although we need only a few inquiries to compute actual ZO gradients, the statistical behavior of a gradient (rather than the value of the individual gradient) decides the training performance. To verify statistical asymptotic behavior, we set $n=1000$ to perform a Monte Carlo computation to get empirical mean values of Q-RGE1 and Q-RGE2, and then compare them with a full-precision ZO gradient via the $\ell_2$ error. As shown in Fig.~\ref{fig:quzo_analysis} (a), the expected values of both quantized estimators have larger errors as the precision reduces from ${\rm INT}8$ to ${\rm INT}3$. However, our method (Q-RGE2) is much more resilient to quantization errors and has a more accurate expected value, since our quantized ZO gradient estimator can avoid the additional bias caused by quantization.  

{\bf Training Behavior.} %Here, we analyze the impact of our proposed RGE method within the QuZO framework. The naive quantized ZO approach approximates the gradient by directly applying \(\hat{\mathbf{u}}_i = Q(\mathbf{u}_i)\), referred to as Q-RGE1 in Eqn.~\ref{equation:mezo_quant}. However, the proposed RGE method, denoted as Q-RGE2, utilizes two conditionally independent quantized vectors, \(\mathbf{u}_{i,1}\) and \(\mathbf{u}_{i,2}\), to estimate the gradient of the loss function \(\mathcal{L}\) with respect to the quantized model parameters \(\bar{\mathbf{w}}\), as formulated in Eqn.~\ref{equation:qzo_approx}. 
Figure~\ref{fig:quzo_analysis} (b) further shows the training behavior of quantized ZO optimization using these two gradient estimators when fine-tuning the OPT-1.3B model. Experiments are performed on the DROP dataset under 8-bit and 4-bit settings. We observe that our QuZO with Q-RGE2 shows slightly better convergence compared to quantized training using Q-RGE1 in the 8-bit setting. In 4-bit training, our method demonstrates a stable and significantly better training behavior: it achieves a loss similar to 8-bit training, while ${\rm INT}$ 4 Q-RGE1 causes convergence failures. 
The above analysis clearly demonstrates the better numerical performance of our QuZO in low-bit LLM fine-tuning.

\begin{figure*}
    \centering
    \includegraphics[width=1\linewidth]{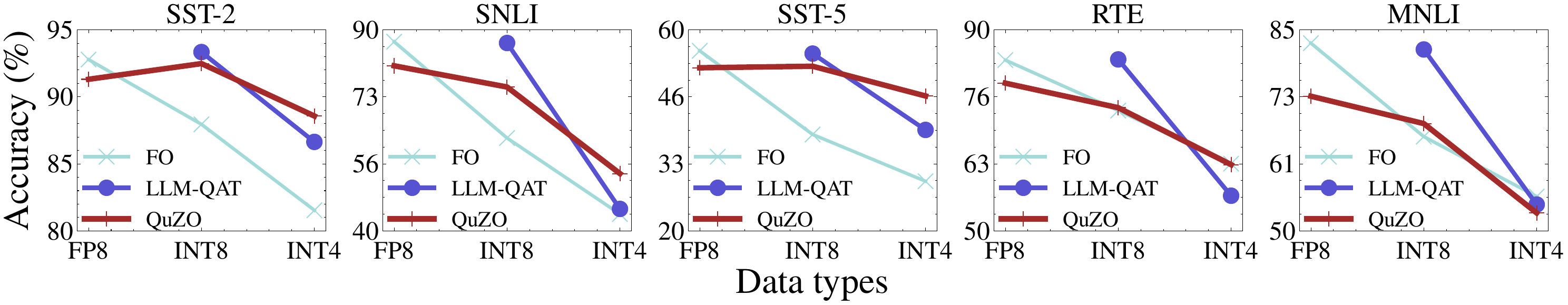}    
    \caption{Experimental findings on RoBERTa-large (350M parameters) with prompts reveal that QuZO, leveraging full-parameter tuning, starts to surpass FO and LLM-QAT as precision reduces to ${\rm INT}8$ or below.}
    \label{fig:roberta_taskwise}
    \vspace{-5pt}
\end{figure*}

\begin{table*}[h]
\centering
\caption{QuZO demonstrates superior performance in full-parameter fine-tuning of LLaMa-2 7B. Note : W\(a\)A\(b\) quantization configurations, which refer to \(a\)-bit weight quantization and \(b\)-bit activation quantization.}
\label{tab:llama-2}
\resizebox{0.85\textwidth}{!}{%
\begin{tabular}{cc|cccc|cc|cc}
\hline
\multicolumn{2}{c|}{LLaMa-2 7B Model} & \multicolumn{4}{c|}{Classification} & \multicolumn{2}{c|}{Multiple-Choise} & \multicolumn{2}{c}{Generation} \\ \hline
\multicolumn{1}{c|}{Data Precision} & Method & \multicolumn{1}{c|}{SST-2} & \multicolumn{1}{c|}{RTE} & \multicolumn{1}{c|}{WSC} & MultiRC & \multicolumn{1}{c|}{COPA} & ReCoRD & \multicolumn{1}{c|}{SQuAD} & DROP \\ \hline
\multicolumn{1}{c|}{FP} & FO & 95.41 & 63.73 & 63.46 & 65.10 & 86.00 & 81.00 & 90.71 & 51.38 \\
\multicolumn{1}{c|}{W32A32} & MeZO & 94.80 & 54.60 & 58.80 & 62.60 & 82.70 & 70.80 & 72.50 & 46.80 \\ \hline
\multicolumn{1}{c|}{FP} & FO & \textbf{91.63} & \textbf{63.90} & 49.00 & \textbf{58.00} & 79.00 & 72.50 & 72.68 & 23.46 \\
\multicolumn{1}{c|}{W8A8} & \textbf{QuZO} & 91.05 & 55.59 & \textbf{65.38} & 57.10 & \textbf{80.00} & \textbf{76.80} & \textbf{76.38} & \textbf{30.17} \\ \hline
\multicolumn{1}{c|}{} & FO & 90.81&  52.34 & 61.53 & 50.60 & 62.00 & 74.83 & 70.13 & 20.06 \\
\multicolumn{1}{c|}{INT} & SmoothQuant & 91.05 & \textbf{66.78} & 59.51 & \textbf{61.50} & 72.02 & 79.10 & 73.07 & 29.94 \\
\multicolumn{1}{c|}{W8A8} & LLM.int8() & 88.04 & 62.56 & 57.75 & 55.61 & 80.02 & \textbf{80.61} & 76.34 & 20.15 \\
\multicolumn{1}{c|}{} & \textbf{QuZO} & \textbf{92.00} & 61.01 & \textbf{63.46} & 60.00 & \textbf{81.00} & 79.00 & \textbf{77.71} & \textbf{30.11} \\ \hline
% \multicolumn{1}{c|}{} & \textbf{QuZO(Sparse)} &  89.56 & 56.67  & 61.53  & 57.86  & 75.00 &  78.00 &  \textbf{81.18} & 22.34  \\ \hline
\multicolumn{1}{c|}{} & FO & 86.35 &  47.29  & 60.57 & 51.90 & 62.04 & 73.21 & 30.01 & 10.06 \\
\multicolumn{1}{c|}{INT/FP} & MinMax & 88.1 & 59.91 & 41.28 & 53.21 & 82.51 & 80.97 & 50.07 & 24.71 \\
\multicolumn{1}{c|}{W4A8} & LLM-FP4 & 91.20 & \textbf{66.82} & 61.38 & 58.81 & \textbf{82.90} & \textbf{81.25} & 51.07 & \textbf{24.99} \\
\multicolumn{1}{c|}{} & \textbf{QuZO} & \textbf{91.62} & 54.87 & \textbf{62.28} & \textbf{60.60} & 80.01 & 78.20 & \textbf{68.12} & 21.80  \\ \hline
% \multicolumn{1}{c|}{} & \textbf{QuZO(Sparse)} & 87.15 & 63.17 & 61.33 & 56.82  & 78.00 &  78.76 &  63.09 &  21.54 \\\hline
\end{tabular}%
}
    \vspace{-10pt}
\end{table*}

% \begin{figure*}
%     \centering
%     \includegraphics[width=1\linewidth]{figure/roberta_taskwise.pdf}
%     \caption{Experimental findings on RoBERTa-large (350M parameters) with prompts reveal that QuZO, leveraging full-parameter tuning, starts to surpass FO and LLM-QAT as precision reduces to ${\rm INT}8$ or below.}
%     \label{fig:roberta_taskwise}
%     \vspace{-5pt}
% \end{figure*}

\section{Experiments}

\begin{table*}[]
\centering
\small
\caption{Results of parameter-efficient fine-tuning of quantized LLMs.}
\label{tab:quzo_lora}
\resizebox{0.9\textwidth}{!}{%
\begin{tabular}{c|c|c|c|c|c|c|c}
\toprule
\textbf{Model} & \textbf{Methods} & \textbf{Gradient} & {\textbf{SST-2}} & \textbf{MultiRC} & \textbf{ReCoRD} & \multicolumn{1}{c|}{\textbf{SQuAD}} & \textbf{DROP} \\ 
\midrule

\multirow{3}{*}{8bit LLaMa2-7B}  
  & FO (LoRA) & ${\rm INT}8$ & 91.97  & 45.60  & 80.10  & 56.13 & 20.93 \\ 
  & MeZO (LoRA) & ${\rm FP}32$ & 88.17 & \textbf{63.60}  & \textbf{80.60} & \textbf{86.96} & \textbf{37.23}  \\
  & \textbf{QuZO (LoRA)} & ${\rm INT}8$ & \textbf{90.36} & 60.00 & 80.80 & 79.97 & 36.11 \\

\midrule

\multirow{3}{*}{8bit OPT-1.3B}  
  & FO (LoRA) & ${\rm INT}8$ & \textbf{91.97} & 55.30 & 70.50 & \textbf{71.92} & 18.35 \\ 
  & MeZO (LoRA) & ${\rm FP}32$ & 89.56 & \textbf{57.30} & \textbf{70.80} & 68.11 & 18.53 \\
  & \textbf{QuZO (LoRA)} & ${\rm INT}8$ & 88.76 & 55.90 & 70.20 & 66.53 & \textbf{23.17} \\

\midrule

\multirow{3}{*}{4bit OPT-1.3B}  
  & FO (LoRA) & ${\rm INT}4$ & 53.89  & 55.55  & 17.20  & 28.20  & 10.00  \\ 
  & MeZO (LoRA) & ${\rm FP}32$ & \textbf{90.48} & 54.10 & \textbf{69.50} & \textbf{70.93} & 20.53 \\
  & \textbf{QuZO (LoRA)} & ${\rm INT}4$ & 87.84 & \textbf{56.20} & 68.90 & 68.13 & \textbf{21.92}  \\

\bottomrule
\end{tabular} }
\end{table*}
In this section, we evaluate the proposed QuZO method on several language models (LMs) with 4-8 bit precision. QuZO demonstrates performance comparable to or better than standard first-order (FO) truly quantized training across various model sizes and tasks, with significantly lower memory usage. We also explore fine-tuning quantized models by combining QLoRA~\citep{dettmers2024qlora} with QuZO. For hardware costs, QuZO employs a forward-only framework with hardware requirements similar to post-training quantization. In Section~\ref{section:mem}, we compare the memory consumption between truly quantized FO training and QuZO. Furthermore, we employ both medium-size models (e.g. RoBERTa-Large~\cite{liu2019roberta})  and large decoder-based LMs [e.g. OPT 1.3B ~\cite{zhang2022opt} and LLaMa-2 7B~\cite{touvron2023llama}] in few-shot settings. All experiments were carried out on NVIDIA A100-40GB GPUs. The details of the experimental setup are provided in Appendix A.

\subsection{Full-Parameter Quantized Fine Tuning} 
We first summarize our experiments on full-parameter fine-tuning for medium- and large-scale models. These results demonstrate that QuZO provides a practical approach for accurate fine-tuning of quantized LLMs directly on low-precision hardware, maintaining. For medium-scale models like RoBERTa-Large, QuZO surpasses truly quantized FO fine-tuning in most tasks in the 4-bit precision. For large-scale models such as LLaMA-2, QuZO achieves performance comparable to or better than truly quantized FO fine-tuning, particularly under ultra-low bit configurations. These findings highlight the ability of QuZO to enable low-cost hardware training without compromising performance. 

\paragraph{Performance on the RoBERTa-Large model.} We evaluate the performance of various methods in the SST-2, SNLI, SST-5, RTE, and MNLI datasets and on the RoBERTa-Large model. The results in Fig.~\ref{fig:roberta_taskwise} leads to the following observations:
\begin{itemize}[leftmargin=*]
\vspace{-5pt}
    \item As expected, all training methods experience accuracy decline as quantization precision decreases. This occurs because the model expressive power declines and the optimization becomes more challenging in lower precision. 
\vspace{-8pt}
\item The performance of truly quantized FO fine-tuning drops most significantly due to the increasing errors in the straight-through estimators as precision decreases.
\vspace{-8pt}
\item Quantization-aware training (QAT) can mitigate partially the accuracy drop of truly quantized FO training. As a faked quantized training, QAT still needs backpropagation and updates LLM model parameters in full precision. Therefore, it remains memory-intensive and less suitable for resource-constrained low-precision hardware. 
\vspace{-8pt}
\item In contrast, the performance of QuZO is {\bf most resilient to the decreased precision}, and it works the best in a very low-precision (e.g., ${\rm INT}4$). This is because (1) QuZO can bypass the error-prone straight-through estimator that is used in truly quantized FO training, and (2) the quantized RGE in Eqn.\eqref{equation:qzo} can eliminate the bias caused by quantized perturbations. 

\end{itemize}

%With the advanced LLM-QAT fine-tuning method, quantized training with STE achieves top performance in the ${\rm INT}8$ setting but performs worse at ${\rm INT}4$ precision.  

%work consistently for smaller LMs during 4/8-bit fine-tuning. Compared with FO fine-tuning, QuZO succeeded in maintaining better model accuracy for W4A8 quantization on 3 out of 5 tasks and also performed better in W8A8 quantization by adopting the INT format in all five tasks. We further compare the results with recent LLM-QAT work, LLM-QAT applied a dynamic quantization with an asymmetric strategy which no doubt improved performance a lot. However, QuZO can be better than the LLM-QAT approach in these 5 tasks for W4A8 quantization. 

\paragraph{Performance of QuZO on LLaMA Models.} We further apply QuZO to fine-tune the LLaMa-2 model, evaluating it on SuperGLUE~\cite{wang2019superglue} and generation tasks. Table~\ref{tab:llama-2} shows that QuZO outperforms its truly quantized FO counterparts on all multichoice and generation tasks under FP W8A8 quantization (i.e. ${\rm FP}8$ for both weights and activations). Under the INT W8A8 quantization, QuZO outperforms SmoothQuant, LLM.int8(), and truly quantized FO methods in 4 out of 8 tasks. For 4-bit quantized FO training, uniform quantization yields the worst accuracy, but advanced methods such as LLM-FP4 improve performance. LLM-FP4~\citep{liu2023llm4} and its baseline MinMax use FP W4A8 quantization and achieve a slight improvement in accuracy, particularly for multichoice tasks. Our QuZO method maintains strong performance under W4A8 quantization with mixed-datatype (see Appendix B), achieving the best results in 4 out of 8 tasks. SmoothQuant, LLM.int8(), MinMax, and LLM-FP4 have introduced efficient quantization methods that enhance performance. However, they are memory-intensive as they require fine-tuning using a FO optimizer. 
% \zz{Are SmoothQuant, LLM.int8(), MinMax and LLM-FP4 first-order training? If yes, please make this clear, and emphasize that they are more memory expensive.}In INT W4A8/W8A8 settings, QuZO uses 8-bit quantized perturbations for FT tuning as shown in Table~\ref{tab:llama-2}.

% \usepackage{graphicx}
\begin{table}[]
\centering
\caption{Performance Comparison of QuZO on the LLaMa-2 13B Model}
\label{tab:llama2_13b}
\resizebox{0.45\textwidth}{!}{%
\begin{tabular}{c|c|c|c|c}
\hline
Model (\#Bit) & \textbf{Methods} & \textbf{ReCoRD} & \textbf{SQuAD} & \textbf{DROP} \\ \hline
              & FO               &       81.70          &     63.23           & 25.90     \\
LLaMa2-13B    & MeZO             &     82.10            &      63.71          &      25.20         \\
(8-Bit)       & QuZO             & \textbf{82.20}            & \textbf{78.19}          & \textbf{37.61}         \\ \hline
              & FO               & 82.00      &          62.27      &       25.31        \\
LLaMa2-13B    & MeZO             &      \textbf{82.30}           &       62.62         &          25.33     \\
(4-Bit)       & QuZO             &     82.10            & \textbf{73.79}          &       \textbf{27.32}        \\ \hline
\end{tabular}%
}
\vspace{-10pt}
\end{table}

\subsection{Parameter-Efficient Fine-Tuning}
\label{subsec: QuZO+LoRA}

Parameter-efficient fine-tuning methods like QLoRA~\citep{dettmers2024qlora} reduce memory usage with 4-bit precision compared to standard training but still rely on AdamW~\citep{loshchilov2017decoupled}, which requires backpropagation. QuZO improves inference efficiency and memory savings, achieving a $5.47 \times$ reduction in maximum memory cost compared to QLoRA in fine-tuning the 4-bit OPT-1.3B model (details in Appendix C). 

Our QuZO framework applies the LoRA (rank set as $8$), allowing fine-tuning with far fewer trainable parameters than full-model tuning, significantly reducing memory consumption, and accelerating convergence. Table~\ref{tab:quzo_lora} highlights the performance of QuZO with low-bit perturbation and gradient configurations for different tasks and models. For the OPT-1.3B model, QuZO utilizes ${\rm INT}8$ RGE gradients with ${\rm INT}4$ perturbations. Despite the introduction of low-bit gradients, QuZO achieves competitive or superior performance compared to full-precision MeZO with LoRA in most tasks and demonstrates strong robustness in 4-bit fine-tuning, while truly quantized FO shows poor accuracy in 4-bit training. Furthermore, QuZO reduces $2-5.47\times$ memory consumption compared to fully quantized FO methods in Table~\ref{tab:memory_comparison}(see Appendix C). For the LLaMa2-7B model, QuZO achieves performance comparable to full-precision MeZO while allowing fine-tuning across all five tasks on a single GPU. In contrast, truly quantized FO methods encounter out-of-memory (OOM) issues. This result highlights that the low-bit stochastic perturbation of QuZO effectively reduces memory overhead while mitigating the bias caused by quantization errors, making accurate fine-tuning feasible on resource-constrained hardware.

\paragraph{Fine-Tuning 13B LMs.} Table 3 presents the performance comparison of QuZO fine-tuning against other methods with LoRA, including First-Order (FO) and MeZO, on the LLaMa-2 13B model under 8-bit and 4-bit quantization. The evaluation is conducted on three datasets: ReCoRD, SQuAD, and DROP, which assess reading comprehension and reasoning ability. The results indicate that QuZO consistently outperforms MeZO and FO, particularly in SQuAD and DROP, demonstrating its ability to better retain performance in a quantized setting. In the 8-bit setting, QuZO achieves a significant improvement. In the 4-bit setting, the trend remains similar, highlighting the robustness of QuZO in handling more aggressive quantization.

\begin{figure}[t]
    \centering
    \includegraphics[width=\linewidth]{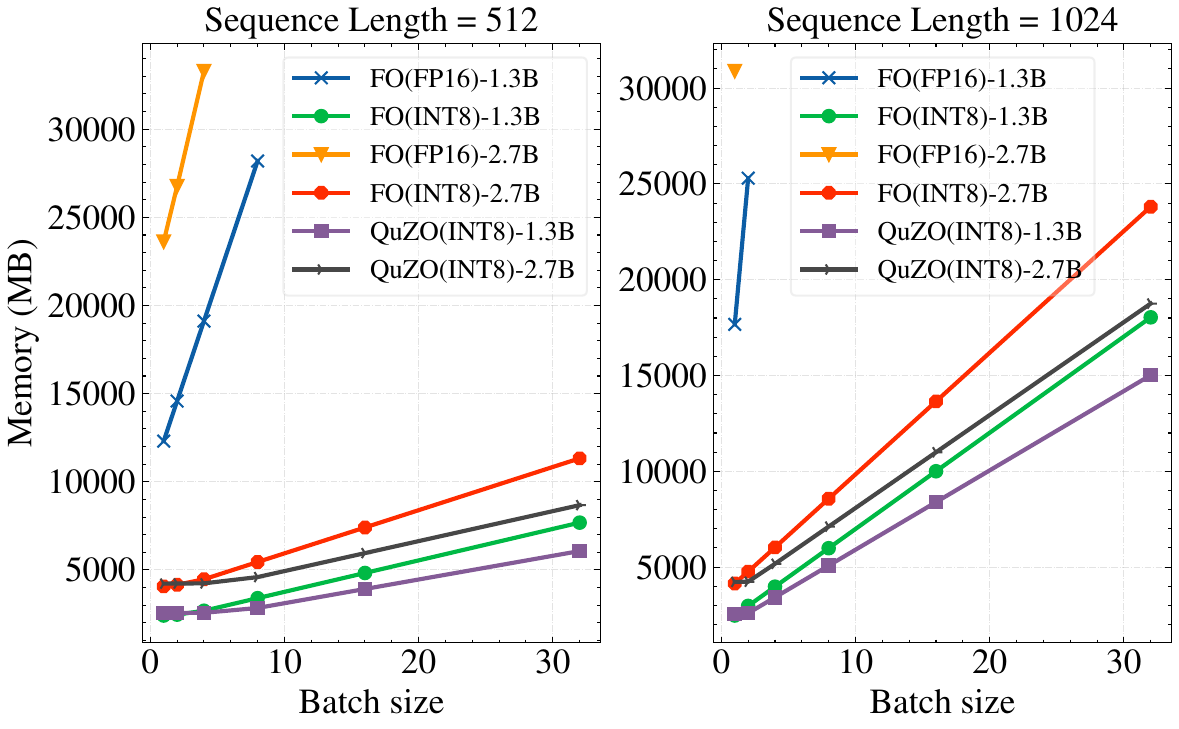}
    \caption{Peak memory usage of FP16 and INT8 training on the OPT 1.3B/2.7B model with sequence lengths of 512 (left) and 1024 (right).}

    \label{fig:hardware_bench}
    \vspace{-5pt}
\end{figure}

\subsection{Memory Efficiency}
\label{section:mem}
We follow Table 5 of \cite{zhang2024revisiting} to provide the theoretical analysis for different optimizers. As shown in Table~\ref{tab:mem_comp}, our QuZO demonstrates significant memory reduction compared to truly quantized FO fine-tuning at the same precision. 

We further compare the empirical memory costs of full fine-tuning the LLaMA-2 7B model in Table~\ref{tab:memory_results}. Specifically, in the MultiRC task, QuZO (8-bit) reduces memory usage by 1.43$\times$, while QuZO (4-bit) achieves a 1.39$\times$ reduction compared to their truly quantized FO counterparts with the same precision. Similarly, in the SQuAD task, QuZO (8-bit) reduces memory consumption by 2.94$\times$, and QuZO (4-bit) achieves a 2.89$\times$ reduction relative to FO-SGD at the same precision. 

To verify hardware efficiency, we profile the memory usage of our QuZO method with ${\rm INT}8$ CUDA kernels, comparing it to the peak memory consumption of ${\rm INT}8$ and ${\rm FP}16$ tensor-core GEMM implementations in full parameter tuning. In practice, QuZO achieves up to a 7.8$\times$ memory reduction with an ${\rm INT}8$ model compared to the first-order ${\rm FP}16$ trainning, as shown in Fig~\ref{fig:hardware_bench}.

\begin{table}[t]
\caption{Comparison of peak memory consumption during full-model fine-tuning. Note: model storage (Weight Mem.) and dynamic allocations for gradients (Dynamic Mem.). $\vert\mat{w}\vert$ and $\vert\mat{a}\vert$ denote memory usage for model parameters and intermediate parameters, respectively, with $l$ representing a specific layer.}
\label{tab:mem_comp}
\resizebox{0.5\textwidth}{!}{%
\begin{tabular}{ccc}
\hline
\multicolumn{1}{c|}{Method} & \multicolumn{1}{c|}{Weight Mem.} & Dynamic Mem \\ \hline
\multicolumn{3}{c}{Full Precision Optimizer}                \\ \hline
\multicolumn{1}{c|}{FO-SGD}         & \multicolumn{1}{c|}{$\vert\mat{w}\vert$} & $\sum_l \max \ \{ \vert\mat{a}\vert, \vert\mat{w}\vert \}$ \\ \hline
\multicolumn{1}{c|}{MeZO}       & \multicolumn{1}{c|}{$\vert\mat{w}\vert$}  & $\max_l{\vert\mat{w}\vert}$ \\ \hline
\multicolumn{3}{c}{Optimizer with Low Precision Model}      \\ \hline
\multicolumn{1}{c|}{FO(8-bit)}   & \multicolumn{1}{c|}{$\vert\mat{w}\vert /4$}  & $\sum_l \max \ \{  \frac{\vert \mat{a}\vert}{4}, \frac{\vert \mat{w}\vert}{4} \}$ \\ \hline
\multicolumn{1}{c|}{FO(4-bit)}   & \multicolumn{1}{c|}{$\vert\mat{w}\vert /8$}  & $\sum_l \max \ \{  \frac{\vert \mat{a}\vert}{8}, \frac{\vert \mat{w}\vert}{8} \}$ \\ \hline
\multicolumn{1}{c|}{QuZO(8-bit)} & \multicolumn{1}{c|}{$\vert\mat{w}\vert /4$}  & $\max_l{\frac{\vert \mat{w}\vert}{4}}$  \\ \hline
\multicolumn{1}{c|}{QuZO(4-bit)} & \multicolumn{1}{c|}{$\vert\mat{w}\vert /8$}  & $\max_l{\frac{\vert \mat{w}\vert}{8}}$ \\ \hline
\end{tabular}%
}
\end{table}

\begin{table}[t]
    \centering
    \small
    \caption{Total memory consumption (GB) for different optimizers on LLaMa-2 7B.}
    \label{tab:memory_results}
    \resizebox{\linewidth}{!}{
    \begin{tabular}{l|cc}
        \hline
        \textbf{Method} & \textbf{MultiRC (GB)} & \textbf{SQuAD (GB)} \\
        \hline
        FO-SGD (8-bit) & 11.66 & 21.29 \\
        FO-SGD (4-bit) & 6.28 & 10.73 \\
        QuZO (8-bit) & 8.15 & 7.24 \\
        QuZO (4-bit) & 4.52 & 3.71 \\
        \hline
    \end{tabular}
    }
\end{table}

\section{Conclusion}

This work has proposed a Quantized Zeroth-Order (QuZO) method for truly qantized training of LLMs without using back propagation. We have identified the challenge of quantized ZO training, and proposed an new quantized ZO gradient to mitigate the bias in low-precision settings. Since QuZO can avoid the error-prone straight-trough estimator, it can achieve better performance than first-order truly quantized training in low-bit settings. The superior performance of QuZO in low-bit (e.g., ${\rm INT}$8 and ${\rm INT}$4) training has been shown by a variety of fine-tuning experiments on the OPT-1.3B and LLaMA2-7B models.  Our QuZO method is intrinsically hardware efficient for fine-tuning LLMs on low-bit resource-constrained hardware.

%%%% Reference %%%%
\section*{Limitations}
The presented QuZO method can significantly impact practical LLM deployment. We have not yet implemented the real quantized training framework using low-precision kernels during training, as this requires much engineering effort. For instance, adding a minimal hardware block to an LLM inference accelerator can enable resource-efficient fine-tuning, making on-device learning of LLMs accessible and affordable for many downstream users. Additionally, QuZO can greatly reduce the latency and energy cost of fine-tuning due to its capability to directly use an ultra low-bit LLM inference accelerator. This will enable the deployment of LLMs in many resource-constrained scenarios, such as autonomous systems and robots.

\bibliography{custom}

% \clearpage
% \bibliographystyle{plainnat}
% % Entries for the entire Anthology, followed by custom entries
% \bibliography{custom}

\clearpage
\appendix
\section*{Appendix}
\label{sec:appendix}

\appendix
% \section{Appendix}

\section{Experiments Setup} 
We first conduct experiments with RoBERTa-large on sentiment classification and natural language classification tasks. We follow prior works~\cite{malladi2024fine} in low data resource settings which can be sampling $k$ examples per class for $k$ = 16 or 512. QuZO is running for 100k steps and the first order fine-tuning for 5 epochs. We also conducted experiments on a smaller set of tasks~\cite{wang2018glue} that includes entailment, span sentiment analysis, and topic classification. These tasks include perceptual analysis (SST-2 and SST-5 \cite{socher2013recursive}), Question Classification (TREC~\cite{hovy2001toward}),, and natural language reasoning (MNLI, SNLI, and RTE \cite{bowman2015large, williams2017broad, rajpurkar2018know}). The metrics we used for the GLUE benchmark are summarized in Table \ref{tab:glue_metric}.
\begin{table}[h]
\centering
\caption{Metrics that we use to evaluate GLUE Benchmark for BERT-based Model.}
\label{tab:glue_metric}
\resizebox{0.25\textwidth}{!}{%
\begin{tabular}{@{}cc@{}}
\toprule
Task Name & Metric                       \\ \midrule
% QNLI      & Accuracy                          \\
SST-2     & Accuracy                      \\
SST-5      & Accuracy          \\
MNLI      & Matched Acc.          \\
SNLI      & Accuracy          \\
% CoLA      & Matthews corr.                       \\
% MRPC      & F1                                    \\
% STS-B     & Spearman corr.                       \\
TREC     & Accuracy                    \\
RTE       & Accuracy                          \\ \bottomrule
% QQP       & F1                                   \\ \bottomrule
\end{tabular}
}
\end{table} ~\\
Subsequently, we selected several SuperGLUE tasks~\cite{wang2019superglue}, encompassing classification (CB, BoolQ, WSC) and multiple-choice (COPA and ReCoRD), alongside two additional question-answering tasks (SQuAD~\cite{rajpurkar2016squad} and DROP~\cite{dua2019drop}). To intensify the challenge, we operated under the few-shot setting, randomly sampling 1,000 examples for training, 500 for validation, and 1,000 for testing. We followed the prompt settings outlined in Appendix D of~\cite{malladi2024fine} to adapt classification tasks into language model tasks. The evaluation metrics used are summarized in Table \ref{tab:superglue_metric}. All experiments were conducted using the AdamW optimizer \cite{loshchilov2018decoupled}.
\begin{table}[h]
\centering
\caption{Metrics that we use to evaluate SuperGLUE and generations tasks.}
\label{tab:superglue_metric}
\resizebox{0.2\textwidth}{!}{%
\begin{tabular}{@{}cc@{}}
\toprule
Task Name & Metric                       \\ \midrule
CB      & F1                          \\
BoolQ     & Accuracy                      \\
WSC      & F1          \\
COPA      & Accuracy                        \\
ReCoRD      & F1                                    \\
SQuAD    & F1                       \\
DROP     & F1                        \\
\bottomrule
\end{tabular}%
}
\vspace{-10pt}
\end{table}
\subsection{Hyperparameters}

\begin{table}[t]
\centering
\caption{The hyperparameter grids used for RoBERTa-Large experiments.}
\label{tab:roberta_hyper}
\resizebox{0.45\textwidth}{!}{
\begin{tabular}{c|c|c}
\toprule
\textbf{Experiment} & \textbf{Hyperparameter} & \textbf{Values} \\ 
\midrule
\multirow{2}{*}{FO}   
           & Batch size      & $\{8,16\}$  \\  
           & Learning rate   & $\{1e{-}5, 1e{-}6\}$  \\ 
\midrule
\multirow{2}{*}{LLM-QAT}   
           & Batch size      & $\{8,16\}$   \\  
           & Learning rate   & $5e{-}6$  \\ 
\midrule
\multirow{4}{*}{QuZO}   
           & Batch size      & $\{16,64\}$   \\  
           & Learning rate   & $\{1e{-}6,1e{-}7\}$   \\ 
           & $\epsilon$      & $1e{-}5$  \\  
           & Weight decay    & $\{0, 0.1\}$ \\  
\bottomrule
\end{tabular}
}
\end{table}
As observed in some LLM fine-tuning literature, zeroth-order (ZO) optimization typically shows consistent performance improvement with training steps. However, the number of forward passes significantly affects computational costs. To optimize resource usage, we limit the training steps to 10k for the RoBERTa-Large model on the SST-2, SST-5, TREC, MNLI, and SNLI datasets. In Table~\ref{tab:roberta_hyper}, our method primarily use a batch size of 64 and experiment with different learning rates for RoBERTa-Large fine-tuning (Fig.~\ref{fig:roberta_taskwise}). Since first-order (FO)-based methods use the Adam optimizer, both FO and LLM-QAT~\citep{liu2023llm} experiments utilize smaller batch sizes and larger learning rates compared to ZO tuning. We use the hyperparameters in Table~\ref{tab:roberta_hyper} for the RoBERTa-Large model. Note that even though we run all experiments for 5 epochs, further learning steps may help to improve the performance of our proposed methods further. \\~\\
Regarding the LLaMa-2 7B model, we use the hyperparameters in Table~\ref{tab:llama_hyper}. We evaluate the model for around 10-12k training steps and directly use the last checkpoint for evaluation. All first-order (FO) quantization training experiments train for 5 epochs and all QuZO experiments use 12K steps. 

\begin{table}[t]
\centering
\caption{The hyperparameter grids used for LLaMA-2 experiments. }
\label{tab:llama_hyper}
\resizebox{0.45\textwidth}{!}{%
\begin{tabular}{@{}ccc@{}}
\toprule
Experiment & Hyperparameters & Values \\ \midrule
QLoRA         & Batch size      &     $[2,4,8,16] $    \\
           & Learning rate   &    $1e-5,5e-6,5e-7$        \\ \midrule
LLM.int8()         & Batch size      &     $[2,4,8,16] $    \\
           & Learning rate   &    $1e-5,5e-6,5e-7$        \\ \midrule
MeZO         & Batch size      &     $[8,16] $    \\
           & Learning rate   &    $1e-4,5e-5,5e-6$        \\ \midrule
QuZO         & Batch size      &    $[4,8,16] $     \\
           & Learning rate   &     $1e-4,5e-5,5e-6$    \\ \bottomrule
\end{tabular}%
}
\vspace{-10pt}
\end{table}

\paragraph{Modeling and implementation} The model and prompt-tuning process follows a structured approach tailored for RoBERTa-large, OPT, and LLaMa-2 models across various tasks. For RoBERTa, a masked language model (MLM) fine-tuning paradigm is used, where prompts incorporate [MASK] tokens that the model learns to predict, with specific label word mappings defining classification outputs. Tasks such as sentiment classification (SST-2, SST-5), topic classification (TREC), and natural language inference (MNLI, SNLI, RTE) utilize template-based prompts adapted from prior works~\cite{gao2020making}. \\
For OPT and LLaMa-2, the tuning process follows GPT-3-style prompting~\citep{mann2020language} and encompasses three task categories: classification, multiple-choice, and question answering (QA). Classification tasks rely on cross entropy loss for label prediction, while multiple-choice and QA tasks utilize teacher forcing to train on correct outputs. During inference, classification and multiple-choice predictions are determined using the average log-likelihood per token, whereas QA responses are generated through greedy decoding. Additionally, in-context learning with 32-shot examples is employed to maintain stable results.\\
For classification tasks, RoBERTa uses linear probing, while OPT and LLaMa employ LM head tuning to refine task-specific representations. This fine-tuning framework ensures consistent evaluation across datasets and models, leveraging structured prompts to enhance adaptability in both low-data and fully supervised settings.

\paragraph{Full Parameter Tuning Performance of QuZO on OPT Models}We further evaluate our method on the OPT-1.3B model. The activation functions of OPT models are generally more sensitive to quantization errors compared to the LLaMA model, posing some challenges for LLM quantization. In Table~\ref{tab:opt}, our QuZO method outperform post-training quantization methods such as QLLM and SmoothQuant in 8 out of 11 tasks under the INT W8A8 quantization. 

\begin{table*}[]
\centering
\caption{Performance comparisons for weights and activations quantization on the OPT-1.3B model.}
\label{tab:opt}
\resizebox{\textwidth}{!}{%
\begin{tabular}{@{}cc|ccccccc|cc|cc@{}}
\toprule
\multicolumn{2}{c|}{OPT-1.3B Model} & \multicolumn{7}{c|}{Classification} & \multicolumn{2}{c|}{Multiple-Choise} & \multicolumn{2}{c}{Generation} \\ \midrule
\multicolumn{1}{c|}{Data Precision} & Method & \multicolumn{1}{c|}{SST-2} & \multicolumn{1}{c|}{RTE} & \multicolumn{1}{c|}{CB} & \multicolumn{1}{c|}{BoolQ} & \multicolumn{1}{c|}{WSC} & \multicolumn{1}{c|}{WIC} & MultiRC & \multicolumn{1}{c|}{COPA} & ReCoRD & \multicolumn{1}{c|}{SQuAD} & DROP \\ \midrule
\multicolumn{1}{c|}{} & QLLM & 82.45 & 55.59 & 66.07 & \textbf{63.00} & 63.46 & 52.35 & \textbf{56.81} & 71.01 & 59.90 & 61.49 & 15.80 \\
\multicolumn{1}{c|}{INT} & LLM.int8 & 53.66 & 53.79 & 41.07 & 46.32 & 42.31 & 58.46 & 45.72 & \textbf{75.00} & 70.22 & 67.14 & 10.33 \\
\multicolumn{1}{c|}{W8A8} & SmoothQuant & 75.01 & 52.34 & 37.51 & 48.20 & 44.23 & 57.83 & 53.41 & 71.03 & 68.81 & 69.42 & 11.22 \\
% \multicolumn{1}{c|}{} & \textbf{QuZO(Ours)} & 89.33 & 55.23 & 64.28 & 62.8 & 59.61 & 54.38 & 55.5 & 73 & 68.5 & 67.28 & 21.5 \\
\multicolumn{1}{c|}{} & \textbf{QuZO(FT)} & \textbf{91.38} & \textbf{55.61} & \textbf{67.85} & 62.30 & \textbf{63.46} & \textbf{60.03} & 55.91 & 74.00 & \textbf{70.81} & \textbf{73.88} & \textbf{21.82} \\ \bottomrule
\end{tabular}%
}
\end{table*}

% \subsection{Dataset}
% For DeBERTa-Base, we consider classification datasets: SST-2~\cite{socher2013recursive},  MNLI~\cite{williams2017broad}, QNLI~\cite{rajpurkar2018know}, CoLA~\cite{warstadt2018neural}, QQP, STS-B, MRPC and RTE~\cite{dagan2005pascal}. 
\section{Quantization Methods}
In this section, we present our weight-activation quantization method. Since per-channel activation quantization is incompatible with efficient GEMM kernels, we employ per-tensor static activation quantization as our coarsest-grained quantization method and per-channel weight quantization as our finer-grained quantization scheme. For post-training quantization (PTQ) methods, we adopt the quantization configuration from SmoothQuant and evaluate their W8A8 quantization under our low data resource setting. Additionally, we reproduce LLM-FP4~\citep{liu2023llm4} using their open-source codebases and evaluate the same tasks within their frameworks, noting that it requires significant time for datatype searching. To ensure a fair comparison, we reduce the calibration size to 8.

\subsection{Weight-only Quantization}
Throughout this work, we focus initially on both weight and activation quantization. This approach can introduce significant quantization errors and lead to accuracy degradation. To address this, we further evaluate weight-only quantization on several tasks, as detailed in Table~\ref{tab:weight-only}. Our findings indicate that weight-only quantization yields better performance compared to combined weight and activation quantization. There are some related work that only do weight quantization for LLMs (i.e GPTQ~\cite{frantar2022gptq}). But it converts the quantized weight to FP16 on the fly during inference and lead to speed up. 
% \usepackage{booktabs}
% \usepackage{graphicx}
\begin{table*}[]
\centering
\caption{Weight-only Quantization experiments conducted on LLaMa-2 7B model.}
\label{tab:weight-only}
\resizebox{\textwidth}{!}{%
\begin{tabular}{@{}cc|ccccc|cc|cc@{}}
\toprule
\multicolumn{2}{c|}{LLaMa-2 7B Model} & \multicolumn{5}{c|}{Classification} & \multicolumn{2}{c|}{Multiple-Choise} & \multicolumn{2}{c}{Generation} \\ \midrule
\multicolumn{1}{c|}{Data Precision} & Method & \multicolumn{1}{c|}{SST-2} & \multicolumn{1}{c|}{RTE} & \multicolumn{1}{c|}{CB} & \multicolumn{1}{c|}{BoolQ} & MultiRC & \multicolumn{1}{c|}{COPA} & ReCoRD & \multicolumn{1}{c|}{SQuAD} & DROP \\ \midrule
\multicolumn{1}{c|}{INT-W4A32} & QuZO(FT) & 92.43 & 60.28 & 60.71 & \textbf{65.50} & 59.60 & 83.00 & 79.00 & \textbf{82.78} & 37.31 \\
\multicolumn{1}{c|}{INT-W8A32} & QuZO(FT) & 92.77  & \textbf{62.81} & \textbf{71.42} & 64.00 & 60.70 & \textbf{83.00} & \textbf{81.00} & 80.93 & 40.25 \\
\multicolumn{1}{c|}{FP-W8A32} & QuZO(FT) & \textbf{93.69}  & 61.37 & 66.07 & 63.72 & \textbf{60.91} & 81.01 & 79.60 & 80.93 & \textbf{37.86} \\ \bottomrule
\end{tabular}%
}
\end{table*}

\subsection{Hybrid Datatype Support}

\paragraph{Mixed Datatypes Support.} Assigning the same low-bit datatype to both weights and activations in QuZO can lead to accuracy degradation due to the limited precision of 4-bit integers compared to floating-point formats, with activation functions being particularly sensitive to quantization errors. While QLoRA introduced the NF4 datatype to mitigate this issue, our QuZO framework takes it a step further by assessing quantization errors~\cite{jung2019learning} for hybrid formats at the same precision. This mixed-datatype fine-tuning in quantized ZO training effectively preserves performance even under 4-bit quantization. Existing works~\cite{liu2023llm,zhou2023dybit} also incorporate this into their quantization strategy but require customized hardware to support the specific datatype. In our quantization algorithm, we use a set of quantization grids \( \mat{b} = \{ b_{1}, b_{2}, \ldots, b_{i} \} \) and apply the quantization operation \( Q_{b}(w) \) to map a full-precision scalar $w$ to a quantized value as follows:

\begin{align} 
    & Q_{b}(\mat{w}) = b_{i}, \quad i = \arg\min_i \big| w - b_{i} \big|. \nonumber
\end{align}
This notation indicates that the parameter \( w \) is quantized to the closest quantization grid point \( b_{i} \). We define the corresponding quantization error as:
\begin{align}
    \mathbb{E}_{b}(\mat{w}) = Q_{b}(\mat{w}) - \mat{w}.
\end{align}
We use the mean squared error (MSE) as the metric to quantify the quantization loss:
\begin{align}
\label{equation:MSE}
    \mathrm{MSE} = \mathbb{E} \big[ ( \mat{w} - Q_{b}(\mat{w}) )^{2} \big].
\end{align}
where $w$ are the FP32 value, and $p(w)$ stands for the probability density function. The neural network weights are a random variable $w\sim p_{w}(w)$. The quantization range is defined between $b_{\text{min}}$ and $b_{\text{max}}$. Our framework selects the data type that minimizes the MSE for each layer and executes the searching algorithm only once before fine-tuning. Based on our data-type search algorithm, we found that INT quantization is more suitable for weight quantization, offering better hardware efficiency. On the other hand, FP quantization is primarily chosen for activation quantization to maintain good accuracy. This quantization selection offers a more accurate QuZO fine-tuning process. \\
Underflow severely impacts low-bit quantization in LLMs~\cite{lee2023enhancing}, associated with rounding zero values that further degrade model performance. Therefore, we propose a hybrid datatype search in Section 4.2 during quantized zeroth-order training, using existing data formats, including integers and floating-points, which are widely used in hardware platforms. We evaluate the LLaMA-2 model using the hybrid datatype detailed in Table~\ref{tab:mixed_dtype}. Through coarse layer-wise datatype selection, QuZO can boost around 1 to 2\% average performance across these 11 tasks in both W4A8 and W8A8 quantization.

\begin{table*}[]
\centering
\caption{Compared to pure-INT or FP quantized zero-order training, our hybrid datatype (INT and FP) searching algorithm boosts accuracy by 1-2\% for most tasks on the LLaMa-2 7B model.}
\label{tab:mixed_dtype}
\resizebox{\textwidth}{!}{%
\begin{tabular}{ccc|ccccccc|cc|cc|c}
\hline
\multicolumn{3}{c|}{LLaMa-2 7B Model} & \multicolumn{7}{c|}{Classification} & \multicolumn{2}{c|}{Multiple-Choise} & \multicolumn{2}{c|}{Generation} & Avg \\ \cline{1-14}
\multicolumn{1}{c|}{Method} & \multicolumn{1}{c|}{Datatype} & Precision & \multicolumn{1}{c|}{SST-2} & \multicolumn{1}{c|}{RTE} & \multicolumn{1}{c|}{CB} & \multicolumn{1}{c|}{BoolQ} & \multicolumn{1}{c|}{WSC} & \multicolumn{1}{c|}{WIC} & MultiRC & \multicolumn{1}{c|}{COPA} & ReCoRD & \multicolumn{1}{c|}{SQuAD} & DROP & Performance \\ \hline
\multicolumn{1}{c|}{\textbf{QuZO(Ours)}} & \multicolumn{1}{c|}{INT} & W4A8 & 89.10 & 54.87 & 62.50 & 66.60 & 64.42 & 57.99 & 60.60 & \textbf{83.00} & 78.20 & 78.12 & 31.80 & 66.10 \\
\multicolumn{1}{c|}{\textbf{QuZO(Ours)}} & \multicolumn{1}{c|}{INT/FP} & W4A8 & 90.59 & 59.92 & 63.71 & 68.40 & 64.50 & \textbf{59.70} & 59.30 & 80.00 & 78.60 & 79.89 & 33.55 & 67.10 \\
\multicolumn{1}{c|}{\textbf{QuZO(Ours)}} & \multicolumn{1}{c|}{INT} & W8A8 & 93.00 & 61.01 & 64.18 & 80.00 & 63.46 & 52.82 & 60.01 & 81.00 & 79.00 & 77.71 & 31.11 & 67.58 \\
\multicolumn{1}{c|}{\textbf{QuZO(Ours)}} & \multicolumn{1}{c|}{INT/FP} & W8A8 & \textbf{93.08} & \textbf{65.95} & \textbf{64.28} & \textbf{81.10} & \textbf{64.57} & 55.17 & \textbf{60.11} & 83.00 & \textbf{79.60} & \textbf{80.74} & \textbf{36.58} & \textbf{69.47} \\ \hline
\end{tabular}%
}
\end{table*}

\subsection{Quantized Perturbation}
We now explore the ZO gradient quantization, which can accelerate model training without compromising convergence. Using a fully quantized I-BERT~\cite{kim2021bert} as an example, we assign low-bit perturbation to update the INT8 model, as shown in Table~\ref{tab:quant_perturb}. The accuracy drop is less than 1\%, but the memory reduction is around 4-16$\times$ for the random perturbation parameters. In the RoBERTa-Large model, we found that 2-bit perturbation performs better, indicating that quantized perturbation does not significantly affect training performance. This is a huge benefit for ZO training since the perturbations are generated and calculated four times for one training step. 
\begin{table}
\caption{Evaluate the impact of low-bit perturbation on QuZO training for SST-2 tasks using different models.}
\scriptsize
\centering
\label{tab:quant_perturb}
\setlength{\tabcolsep}{1mm}{
\begin{tabular}{@{}c|c|c|c@{}}
\toprule
Model & Model Precision & \begin{tabular}[c]{@{}c@{}}Perturbation\\ (\#bit)\end{tabular} & Performance \\ \midrule
I-BERT        & INT W8A8 & 8 & 92.77 \\
I-BERT        & INT W8A8 & 4 & 92.48 \\
I-BERT        & INT W8A8 & 2 & 91.89 \\ \midrule
RoBERTa-Large & INT W8A8 & 8 & 92.48 \\
RoBERTa-Large & INT W8A8 & 4 & 91.51 \\
RoBERTa-Large & INT W8A8 & 2 & 93.07 \\ \midrule
LLaMa-2 7B & INT W4A8 & 8 & 91.32 \\ \bottomrule
\end{tabular}%
}
\vspace{-10pt}
\end{table}
Current works only focus on sparse parameter perturbations~\cite{liu2024sparse} for reducing gradient estimation variance in RGE. It introduces the masks and applies them to weight perturbations per step. However, we now consider on hardware-efficient side and use low-precision weight perturbation to do ZO gradient estimation in LLM fine-tuning. We further analyze the memory costs of the perturbation parameters $\textbf{u}\in \mathbb{R}^{d}$. At each step, QuZO reuses $\textbf{u}$ four times in Algorithm 1. We evaluated the quantized perturbation experiments on the RoBERTa-Large model, and it costs around 1.63 GB of memory to store each $\textbf{u}$ during one step. However, quantized perturbation would only cost 110 to 410 MB if we quantize it to 2-bit or 8-bit, respectively, based on our estimated results in the Table~\ref{tab:storage-qzo}. Since these results are estimated based on the number of perturbations and storage datatype, real hardware implementation is required to demonstrate the full advantage. We will address this in future work.
\begin{table}[]
\centering
\caption{Number of Perturbations \(\textbf{u}\) for Different Models at Each Step.}
\label{tab:storage-qzo}
\resizebox{0.6\columnwidth}{!}{%
\begin{tabular}{@{}cc@{}}
\toprule
 Model & Number of Perturbation \\ \midrule
 I-BERT    &           406885467       \\
 RoBERTa-Large      &     407935067       \\ \bottomrule
\end{tabular}%
}
\vspace{-8pt}
\end{table}

\paragraph{Handling outliers.} The outliers mainly occur in the activations of transformers and can severely degrade quantization performance if not addressed efficiently~\cite{liu2023llm,liu2023qllm,lin2023awq}. To simplify the quantization process without introducing overhead, we propose an outlier detector that can distinguish outliers from normal values. Our outlier detector can automatically select the outlier threshold to determine a suitable ratio $\alpha$ (Outliers/All data), which is normally around 1\%. We quantize the normal data using a pre-defined quantization datatype and quantize the outlier data using the same precision FP type. As a signed INT8 quantization example, we designate the binary code \(10000000_2\) as an outlier label to identify outlier values in the selected tensor array. Consequently, the valid data range becomes \([-127, 127]\), and we utilize an 8-bit floating-point scheme with adaptive biased bits to efficiently quantize these outlier values. It enables efficient quantization of LLMs across various hardware platforms such as CPU and FPGAs using the QuZO method. 

\paragraph{Loss Landscape.} The effectiveness of ZO fine-tuning for LLMs arises from starting near the optimal loss region. Theoretical analysis in \cite{malladi2024fine} [Lemma 3] links ZO convergence to the low effective rank of Hessian matrix. In quantized training, the Lipschitz smoothness constant \( L \) significantly impacts performance \cite{frumkin2023jumping}. Fig.~\ref{fig:loss_landscape} (See Appendix B) demonstrates the stability of the smoothness of loss function across weight and activation quantization levels, underscoring the effectiveness in low-bit ZO training.
\begin{figure}
    \centering
    \includegraphics[width=\linewidth]{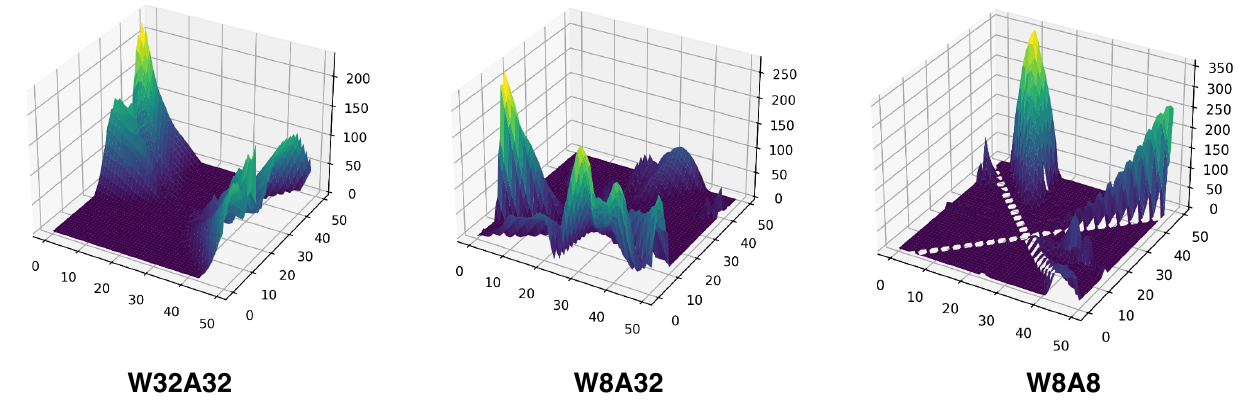}
    \caption{The loss landscape of the RoBERTa-large model under different quantization bits. The notations W and A mean the bits for weights and activation.}
    \label{fig:loss_landscape}
    \vspace{-18pt}
\end{figure}
\subsection{ZO Gradient Accumulation}
Gradient accumulation is a technique for training models where data samples are divided into several batches and calculated sequentially. To fine-tune large models on a single GPU, especially for datasets like DROP that require small batch sizes, we implemented a zeroth-order accumulation method for performing weight updates. Initially, we calculate the gradient without updating the network parameters at each step, accumulating the projected gradient information. After reaching the predefined accumulation steps, the accumulated gradient is used to update the parameters. We also incorporate prevalent efficiency-enhancing tricks adopted in current zeroth-order optimizers, following the first-order approach to implement our zeroth-order method effectively. This approach allows efficient fine-tuning of large models on a single GPU, leveraging the advantages of gradient accumulation within a QuZO optimization framework.

\section{Memory Efficiency of QuZO} To demonstrate the hardware efficiency of QuZO, we employ the Cutlass INT8 Kernel to showcase memory efficiency. To fine-tune large models efficiently with limited GPUs, we assess the first-order (FO) method using Fully Sharded Data Parallelism (FSDP)~\cite{zhao2023pytorch} for distributed training. Besides, We believe it can be further reduced if we fully apply the INT engine in each linear and non-linear layer. This could be our next step in the CUDA optimization.  Finally, we provide the memory cost of our QuZO method using INT8 CUDA kernels and compare it with the peak memory usage of INT8 and FP16 tensor-core GEMM implementations on full parameter tuning. As the batch size increases from 1 to 32, the memory reduction reaches up to 7.8$\times$ when running with an INT8 model compared to FP16 training in Fig. \ref{fig:hardware_bench}. 
\begin{table}[ht]
\centering
\caption{Memory Consumption (GB) Across Models and Methods for Five Tasks. This table compares the memory requirements of different methods (e.g., LLM.int8, QuZO, and QLoRA) across various tasks using two models: OPT1.3B and LLaMa-2 7B. The QuZO method demonstrates significantly lower memory consumption across all models, while LLM.int8() encounters Out of Memory (OOM) issues in some cases.}
\label{tab:memory_comparison}
\resizebox{0.48\textwidth}{!}{%
\begin{tabular}{llccccc}
\toprule
\textbf{Model}            & \textbf{Methods} & \textbf{SST-2} & \textbf{MultiRC} & \textbf{ReCoRD} & \textbf{SQuAD} & \textbf{DROP} \\ \midrule
\multirow{2}{*}{\textbf{8-bit OPT 1.3B}} 
& LLM.int8()  & 9.01   & 23.97  & 6.76   & 22.09  & 31.29  \\
& QuZO      & 3.43   & 12.61  & 4.82   & 7.50   & 16.42  \\ \midrule

\multirow{2}{*}{\textbf{4-bit OPT 1.3B}} 
& QLoRA     & 4.76   & 18.15  & 4.42   & 20.48  & 27.23  \\
& QuZO      & 1.72   & 6.30   & 2.41   & 3.74   & 11.70  \\ \midrule

\multirow{2}{*}{\textbf{8-bit LLaMa-2 7B}} 
& LLM.int8()  & 31.47  & OOM    & 19.06  & OOM    & OOM    \\
& QuZO      & 9.94   & 25.11  & 13.04  & 16.69  & 31.66  \\ \bottomrule
\end{tabular}%
}

\end{table}
Table~\ref{tab:memory_comparison} provides a comprehensive comparison of memory consumption (in GB) across various tasks when fine-tuning quantized models using QuZO with LoRA $(rank=8)$. The methods compared include QuZO, LLM.int8(), and QLoRA. Notably, QuZO employs 4-bit perturbations to fine-tune the models, achieving significant memory savings compared to LLM.int8 and QLoRA. For instance, in the OPT1.3B-int4 model, QuZO reduces memory usage by approximately 2.8$\times$ on SST-2 (1.72 GB vs. 4.76 GB in QLoRA) and by 5.47$\times$ on SQuAD (3.74 GB vs. 20.48 GB in QLoRA). Similarly, for the OPT1.3B-int8 model, QuZO achieves a memory reduction of 1.4$\times$ on MultiRC (12.61 GB vs. 23.97 GB in INT8 FO fine tuning). \\
In the 8-bit LLaMa-2 7B model, while LLM.int8 encounters Out-of-Memory (OOM) errors on several tasks, QuZO successfully completes fine-tuning with substantial memory efficiency, using just 9.94 GB on SST-2 compared to 31.47 GB for LLM.int8—a reduction of 3.2$\times$. These results highlight the ability of QuZO to fine-tune quantized models effectively with minimal memory overhead, leveraging 4-bit perturbations for substantial efficiency gains while maintaining compatibility with LoRA architectures. This positions QuZO as a practical choice for resource-constrained fine-tuning in large-scale NLP tasks.

\end{document}